\documentclass[letterpaper, 10 pt, conference]{ieeeconf}  

\IEEEoverridecommandlockouts                              

\usepackage{balance}
\usepackage{color}

\usepackage{graphics} 
\usepackage{amsmath,amssymb,amsfonts} 
\usepackage{bm}
\usepackage{subcaption}	 
\usepackage{graphicx}
\usepackage{multirow}
\usepackage{array}

\usepackage{url}
\usepackage{hyperref}
\usepackage{xcolor}
\usepackage{}

\usepackage{float}
\usepackage{booktabs}
\usepackage{multirow}
\usepackage{textcomp}
\usepackage{setspace}
\usepackage{tabularx}
\usepackage{makecell}
\usepackage{colortbl}  
\usepackage{pifont}
\newcommand{\cmark}{\ding{51}}
\newcommand{\xmark}{\ding{55}}

\newcommand{\PAR}[1]{\vskip4pt \noindent{\bf #1~}}

\newcommand{\methodname}{L2COcc}

\definecolor{road}{RGB}{255, 0, 255}
\definecolor{sidewalk}{RGB}{75, 0, 75}
\definecolor{parking}{RGB}{255,150,255}
\definecolor{other-grnd.}{RGB}{175,0,75}
\definecolor{building}{RGB}{255,200,0}
\definecolor{car}{RGB}{100,150,245}
\definecolor{truck}{RGB}{80,30,180}
\definecolor{bicycle}{RGB}{100,230,245}
\definecolor{motorcycle}{RGB}{30,60,150}
\definecolor{other-veh.}{RGB}{0,0,255}
\definecolor{vegetation}{RGB}{0,175,0}
\definecolor{trunk}{RGB}{135,60,0}
\definecolor{terrain}{RGB}{150,240,80}
\definecolor{person}{RGB}{255,30,30}
\definecolor{bicyclist}{RGB}{255,40,200}
\definecolor{motorcyclist}{RGB}{150,30,90}
\definecolor{fence}{RGB}{255,120,50}
\definecolor{pole}{RGB}{255,240,150}
\definecolor{traffic-sign}{RGB}{255,0,0}
\definecolor{other-obj.}{RGB}{50,250,250}
\definecolor{other-struct.}{RGB}{255,150,0}

\usepackage[backend=bibtex,natbib=true,style=numeric-comp,sorting=none,giveninits=true,maxbibnames=99,url=false,doi=false]{biblatex}
\title{\LARGE \bf
{\methodname}: Lightweight Camera-Centric Semantic Scene Completion \\ via Distillation of LiDAR Model 
}
\author{Ruoyu Wang$^{1}$, Yukai Ma$^{1}$, Yi Yao$^{1}$, Sheng Tao$^{1}$, Haoang Li$^{2}$, Zongzhi Zhu$^{3}$, Yong Liu$^{1,*}$, Xingxing Zuo$^{2,*}$%
\thanks{$^{1}$The authors are with the Institute of Cyber-Systems and Control, Zhejiang University, Hangzhou, China.}%
\thanks{$^{2}$ The authors are with the School of Computation, Information and Technology,
Technical University of Munich, Munich, Germany.}%
\thanks{$^{3}$ The author is with Zhejiang Guoli Security Technology Co., Ltd.,
Ningbo, China.}%
\thanks{Yong Liu is with the State Key Laboratory of Industrial Control Technology.}
\thanks{$^*$ Xingxing Zuo and Yong Liu are the corresponding authors (Email: {\tt\small xingxing.zuo@tum.de; yongliu@iipc.zju.edu.cn}).}
}
\begin{document}

\maketitle
\thispagestyle{empty}
\pagestyle{empty}

\begin{abstract}
Semantic Scene Completion (SSC) constitutes a pivotal element in autonomous driving perception systems, tasked with inferring the 3D semantic occupancy of a scene from sensory data.
To improve accuracy, prior research has implemented various computationally demanding and memory-intensive 3D operations, imposing significant computational requirements on the platform during training and testing.
This paper proposes {\methodname}, a lightweight camera-centric SSC framework that also accommodates LiDAR inputs.
With our proposed efficient voxel transformer (EVT) and cross-modal knowledge modules, including feature similarity distillation (FSD), TPV distillation (TPVD) and prediction alignment distillation (PAD), our method substantially reduce computational burden while maintaining high accuracy.
The experimental evaluations demonstrate that our proposed method surpasses the current state-of-the-art vision-based SSC methods regarding accuracy on both the SemanticKITTI and SSCBench-KITTI-360 benchmarks, respectively. Additionally, our method is more lightweight, exhibiting a reduction in both memory consumption and inference time by over 23\% compared to the current state-of-the-arts method.
Code is available at our project page: \url{https://studyingfufu.github.io/L2COcc/}.
\end{abstract}
\vspace{-3pt}

\section{Introduction}
Semantic Sene completion (SSC) inferring the semantic occupancy from sensor observations plays a pivotal role in autonomous driving, facilitating obstacle avoidance, path planning, and decision making. 
Following the seminal work by SSCNet~\cite{song2017semantic}, which provided a clear delineation of this task, numerous LiDAR-based methods~\cite{cao2023pasco,cheng2021s3cnet,xia2023scpnet,zuo2023pointocc} have been introduced, leveraging the inherent advantage of LiDAR sensors in capturing accurate geometry information of the scene. However, due to the high cost and installation limitations associated with LiDAR, researchers have shifted their focus toward visual SSC~\cite{cao2022monoscene,zheng2024monoocc,li2023voxformer,mei2023camera}. Likewise, this paper also emphasizes camera-centric SSC.

A critical step in visual SSC is transforming features from 2D images to 3D space. Lift, Splat, and Shoot (LSS)~\cite{philion2020lift} and Deformable Cross Attention (DCA)~\cite{Zhu_Su_Lu_Li_Wang_Dai_2020,li2022bevformer} represent two foundational techniques for feature transformation to generate voxel features. However, DCA-based methods face limitations due to the numerous parameters associated with learnable voxel queries and the substantial memory consumption resulting from three-dimensional deformable self-attention (3D-DSA). Consequently, there is a pressing need for an efficient converter to transform 2D features into 3D voxel features.

%


Moreover, directly predicting semantic occupancy from voxel features produced by the view transformation module can lead to unsatisfying accuracy. Additional refinement steps are required for optimal performance, including the integration of an additional 3D convolutional network~\cite{zhang2023occformer, yu2024context}, as well as multiple stages of DCA and 3D-DSA operations~\cite{jiang2023symphonize, zheng2024monoocc}. However, these supplementary steps can lead to considerable computational and memory overhead, thereby hindering the efficiency and broad applicability of visual SSC.


\begin{figure}[t]
      \centering
      \includegraphics[width=\linewidth]{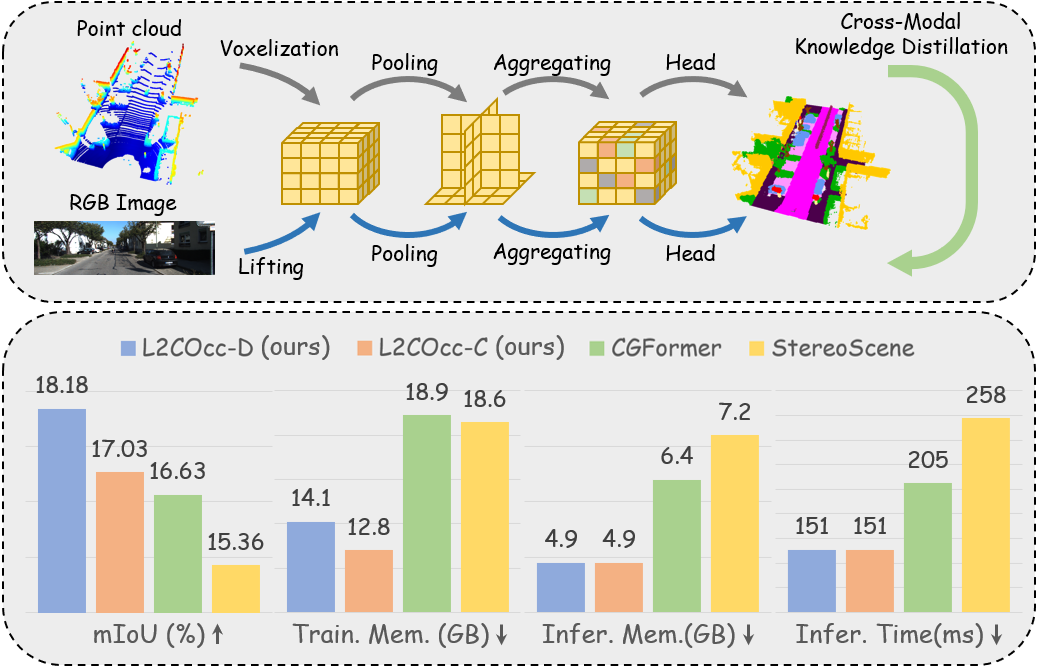}
      \captionsetup{font=small}
      \caption{Top: An overview of our framework. Bottom: Performance of our approach on SemanticKITTI~\cite{behley2019semantickitti} dataset. Our proposed {\methodname} is a TPV-based lightweight camera-centric SSC framework.
      {\methodname-C} is trained using only camera data, whereas {\methodname-D} is distilled from our LiDAR model but operates solely on camera input during inference. Our approach demonstrates superior performance in terms of accuracy, memory consumption, and efficiency compared to representative methods, CGFormer~\cite{yu2024context} and StereoScene~\cite{li2023stereoscene}.}
      \label{fig:cover}
      \vspace{-20pt} 
\end{figure}

Building on the identified challenges above, we propose further advancing visual SSC in this paper by reducing GPU memory usage and improving computational efficiency without compromising accuracy. 
Specifically, we conduct a TPV-based occupancy prediction network (see Fig.~\ref{fig:cover}) that infers semantic occupancy without acquiring complex operations on 3D voxel features. Additionally, we incorporate LiDAR point cloud into the training phase, facilitating knowledge distillation to a camera-only pipeline for enhanced performance at the inference stage. 

The primary contributions of this work can be summarized as follows: 
\begin{itemize}
\item We introduce {\methodname}, a lightweight, TPV-based camera-centric framework for SSC that also accommodates LiDAR input. An efficient voxel transformer (EVT) is proposed, which significantly reduces both the number of parameters and GPU memory usage.
\item We incorporate three types of cross-modal knowledge distillation modules (FSD, TPVD, PAD) applicable to the framework above, utilizing a LiDAR-based teacher to enhance the performance of a camera-based student.
\item  Experimental results show that our {\methodname} outperforms all of the existing visual approaches
on both the KITTI360~\cite{li2023sscbench} and SemanticKITTI~\cite{behley2019semantickitti} benchmarks regarding Intersection over Union (IoU) and mean Intersection over Union (mIoU), despite our method having a significant reduction over 23\% in both memory consumption and inference time compared to the current state-of-the-arts method.
\end{itemize}
%
\section{RELATED WORK}
\label{sec:related work}

\subsection{SSC from LiDAR}
SSC aims to infer semantic occupancy from sensor observation, as defined by SSCNet~\cite{song2017semantic}. 
LiDAR sensors, due to their advantage in accurate 3D geometry measurement, have been extensively used in SSC tasks. SSCNet~\cite{song2017semantic} employs 3D U-Net to voxels and suffers from high computation cost. In contrast, LMSCNet~\cite{roldao2020lmscnet} adopts lighter 2D convolutions, whereas PASCO~\cite{cao2023pasco} employs a transformer-based sparse 3D U-Net to process 3D voxel grid. S3CNet~\cite{cheng2021s3cnet} and SSC-RS~\cite{mei2023ssc} integrate BEV representation with voxel representation to aggregate global information. PointOcc~\cite{zuo2023pointocc} entirely replaces voxel representation with cylinder representation, resulting in a significantly more lightweight and efficient network.

\subsection{SSC from Camera}
The camera is a popular sensor for perception due to its rich semantic information and low economic cost. Pioneering work such as Monoscene~\cite{cao2022monoscene} has positioned monocular visual SSC as a new point of interest. A key step in monocular visual SSC involves generating 3D descriptions from 2D observation. Some methods use DCA-based~\cite{Zhu_Su_Lu_Li_Wang_Dai_2020} approaches to aggregate features from images. TPVFormer~\cite{huang2023tri} aggregates TPV features from images as an alternative to BEV. Voxformer~\cite{li2023voxformer} and Symphonize~\cite{jiang2023symphonize} enhance BEV queries by proposing depth-based and instance queries, respectively. Differently, several methods~\cite{zhang2023occformer,li2023fb,yu2024context,mei2023camera} use LSS view transformer approaches to lift 2D feature to 3D voxels. CGFormer~\cite{yu2024context} fused LSS volume into voxel queries and extended the DCA operation from 2D to 3D pixel space, improving the performance of feature lifting.
Our \methodname utilizes an efficient voxel transformer (EVT) module to extract voxel features, followed by a lightweight TPV-based pipeline to infer semantic occupancy. As shown in Figure~\ref{fig:cover}, {\methodname} is more lightweight and efficient than compared methods, while still maintaining high accuracy.

\begin{figure*}[t]
      \centering
      \includegraphics[width=\linewidth]{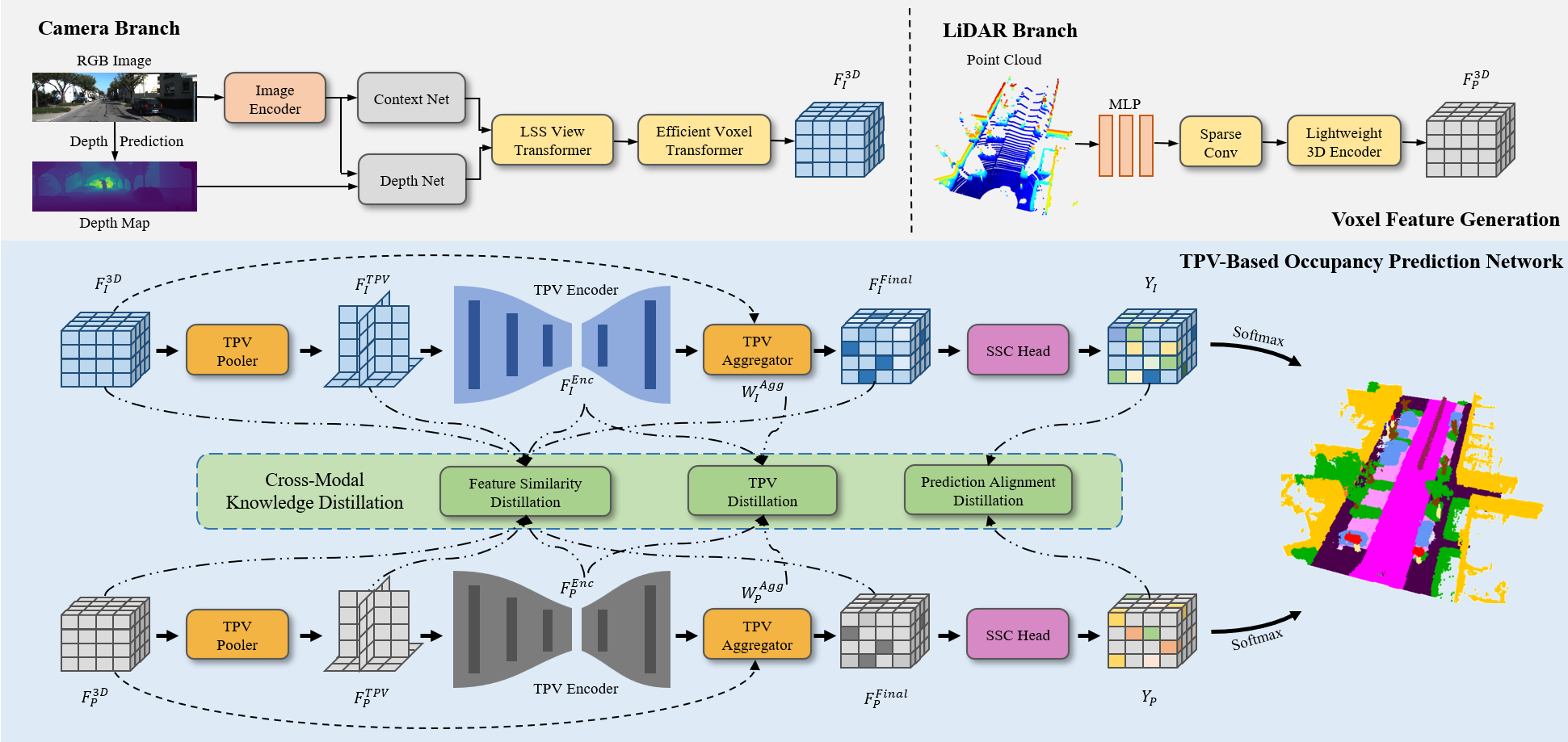}
      \vspace{-1pt}
      \captionsetup{font={small}}
      \caption{The overall framework of our proposed {\methodname}, comprised of three stages: voxel feature generation (indicated by the {\color{gray}gray background}), TPV-based occupancy prediction network (indicated by the {\color{blue}blue background}), and cross-modal knowledge distillation (indicated by the {\color{vegetation}green background}).
      The voxel features generated by the camera and LiDAR branches serve as inputs to the TPV-based occupancy prediction network. Meanwhile, cross-modal knowledge distillation transfers knowledge from the LiDAR-based teacher to the camera-based student, further improving the performance of the camera-only model during inference.
      }
      \label{fig:overrall}
      \vspace{-10pt} 
\end{figure*}

\subsection{Knowledge Distillation}
Knowledge distillation initially focused on the transfer of knowledge between models of different specifications in image classification tasks~\cite{Hinton_Vinyals_Dean_2015}, to enable smaller models to achieve high accuracy. Recent research has shown that knowledge distillation is a general technique that can be applied to a variety of tasks, including BEV object detection~\cite{Zhou_Liu_Hu_Zhou_Ma_2023}, 3D object detection~\cite{Wang_Li_Luo_Xie_Yang_2023,Hong_Dai_Ding}, and also semantic scene completion~\cite{zheng2024monoocc,xia2023scpnet,ma2024licroccteachradaraccurate}. 

SCPNet~\cite{xia2023scpnet} and MonoOcc~\cite{zheng2024monoocc} are  LiDAR- and camera-based methods respectively that both utilize multi-frame teachers to improve the performance of single-frame students. 
LiCROcc~\cite{ma2024licroccteachradaraccurate} proposes a cross-modal distillation method for multi-sensor fusion, which transfers knowledge from the LiDAR-camera fusion model to the radar-camera fusion model, further improving radar-centric semantic occupancy prediction. 

\section{METHODOLOGY}
\label{sec:methodology}

Our objective is to efficiently predict the semantic occupancy of a scene from a single-frame monocular image with a lightweight model. 
Figure~\ref{fig:overrall} shows an overview of our approach. Firstly, we leverage feature extractors to extract 3D voxel features (Section~\ref{voxelfeature}) for images and LiDAR point clouds, respectively. Then, the voxel features are converted into TPV features, and processed by a TPV-based semantic occupancy prediction pipeline (Section~\ref{occnet}). Finally, we transfer knowledge from the LiDAR-based teacher to the camera-based student through knowledge distillation (Section~\ref{KD}), further improving the performance of the camera-only model at inference.

\subsection{Efficient Voxel Feature Extraction}
\label{voxelfeature}


\PAR{Camera Branch.} Given an image $\boldsymbol{\mathbf{I}} \in \mathbb{R}^{H\times W\times3}$, we initially extract 2D image feature map $\boldsymbol{F}^{2D}_{I} \in \mathbb{R}^{H \times W \times C}$ using an image encoder, EfficientNet~\cite{tan2019efficientnet}. 
Subsequently, we transfer the feature $\boldsymbol{F}^{2D}_{I}$ to 3D voxel features  $\boldsymbol{F}^{3D}_{I}$  through the following two steps.

\subsubsection{LSS View Transformation} 
We follow the standard LSS view transformation~\cite{philion2020lift, yu2024context} from 2D into 3D.
Starting from the 2D features $\boldsymbol{F}^{2D}_{I}$ and the predicted depth $\boldsymbol{Z}_{I} \in \mathbb{R}^{H\times W}$ from off-the-self depth prediction model~\cite{bhat2021adabins, shamsafar2022mobilestereonet}, we utilize the Depth Net and Context Net modules proposed in CGFormer~\cite{yu2024context} to generate a discrete depth probability map $\boldsymbol{D} \in \mathbb{R}^{H\times W\times D}$ and context feature $\boldsymbol{C} \in \mathbb{R}^{H\times W\times C}$, respectively. Subsequently, the outer product of $\boldsymbol{D}$ and $\boldsymbol{C}$ is computed to obtain the LSS feature volume, formulated as $\boldsymbol{F}^{LSS}_{I}=\boldsymbol{D} \otimes \boldsymbol{C}$. We supervise the estimated depth probability from the Depth Net by computing the loss $\mathcal{L}_{d}$~\cite{li2022bevformer} with the ground truth provided by LiDAR scans.

\subsubsection{Efficient Voxel Transformer (EVT)}
We back-project the depth map into a point cloud to determine each voxel's occupancy. The occupancy is represented by $\mathbf{M} \in \mathbb{R}^{X \times Y \times Z}$, where voxels occupied by points will be assigned ones otherwise zeros. The LSS feature volume is further selectively processed by 3D deformable cross-attention to obtain $\boldsymbol{F}_I^{EVT}$: 
\begin{align}
    \boldsymbol{F}_I^{EVT} =
    \begin{cases} 
    \text{3D-DCA}(\boldsymbol{F}^{LSS}_I(\boldsymbol{p}), \boldsymbol{F}^{LSS}_I, \boldsymbol{p}), \text{if} \  \mathbf{M}(\boldsymbol{p}) = 1 \\
    \text{MLP}(\boldsymbol{F}^{LSS}_I(\boldsymbol{p})), \ \ \ \ \ \ \ \ \ \ \ \ \  \text{if} \  \mathbf{M}(\boldsymbol{p}) = 0
    \end{cases}
\end{align}
In the above equations, DCA with LSS volume will be employed to aggregate high-level features if the voxel at position $\boldsymbol{p}$ is occupied. Otherwise, a lightweight MLP is used to efficiently generate features.
In the DCA operation, we directly utilize the LSS volume feature as the voxel query rather than using a predefined voxel embedding as existing methods~\cite{li2023voxformer,jiang2023symphonize,zheng2024monoocc,yu2024context}. Sequentially, we employ a lightweight 3D ResNet and FPN (feature pyramid network) to further refine $\boldsymbol{F}_I^{EVT}$ and get the final voxel feature $\boldsymbol{F}^{3D}_{I}$.

\PAR{LiDAR Branch.} Given a LiDAR point cloud $\boldsymbol{P} \in \mathbb{R}^{N \times 3}$, we follow a standard way~\cite{zuo2023pointocc} to get encoded features $\boldsymbol{F}_P \in \mathbb{R}^{N \times C}$, by employing a point-wise MLP and sparse convolution. We then structuralize and densify the encoded sparse features into voxel representation. Same as the camera branch, we also utilize the lightweight 3D ResNet and FPN to obtain LiDAR voxel feature $\boldsymbol{F}^{3D}_{P}$.

\subsection{TPV-Based Lightweight Occupancy Prediction Network}
\label{occnet}
We propose a lightweight TPV-based network to predict semantic occupancy from the 3D voxel feature $\boldsymbol{F}^{3D}$.  $\boldsymbol{F}^{3D} = \boldsymbol{F}^{3D}_{I}  \; \rm{or}  \;  \boldsymbol{F}^{3D}_{P}$. Our network mainly consists of four modules: (1) TPV pooler to compress the 3D voxel feature   $\boldsymbol{F}^{3D}$ to TPV features,  (2) TPV encoder to refine TPV features, (3) TPV aggregator to integrate TPV features back to the voxel feature, and (4) SSC head to infer the semantic occupancy. 

\PAR{TPV Pooler.} Maxpooling-based pooler is straightforward for compression, however, it inevitably incurs information loss. For effective compression,  we propose a weighted average pooling operation (WAP). From the voxel features $\boldsymbol{F}^{3D}$, we compute a weight tensor $\boldsymbol{W} \in \mathbb{R}^{X \times Y \times Z \times 3}$ using a convolution layer ended with softmax operation.
Denote the target dimemsion to be compressed as $d \in \{ 0,1,2\}$, WAP is formulated as:
\begin{align}
    \text{WAP}(\boldsymbol{F}, \boldsymbol{W}, d) = \rm{SUM}((\boldsymbol{W}[:,:,:,d])\odot \boldsymbol{F}^{3D}, d).
\end{align}
The function $\rm{SUM}(\boldsymbol{F},d)$ denotes the operation of summing the tensor $\boldsymbol{F}$ along dimension $d$. We apply pooling operation on three dimensions to obtain three 2D feature maps $\boldsymbol{F}_{xy} \in \mathbb{R}^{X \times Y \times C}$, $\boldsymbol{F}_{yz} \in \mathbb{R}^{Y \times Z \times C}$ and $\boldsymbol{F}_{zx} \in \mathbb{R}^{X \times Z \times C}$.
%

\PAR{TPV Encoder.} To refine the compressed TPV features, we follow CGfomer~\cite{yu2024context} and employ a SwinTranformer-based 2D backbone~\cite{liu2021swin} to extract multi-level features and use a 2D convolutional network to fuse them. We denote the processed TPV features as $\boldsymbol{F}^{Enc} = \{ \hat{\boldsymbol{F}}_{xy}, \hat{\boldsymbol{F}}_{yz}, \hat{\boldsymbol{F}}_{zx}\}$.

\PAR{TPV Aggregator.} TPV features are initially broadcast onto the 3D space while preserving the same shape as the 3D voxel features $\boldsymbol{F}^{3D}$ to facilitate aggregation operations. For example,  $\hat{\boldsymbol{F}}_{xy} \in \mathbb{R}^{X \times Y \times C}$ is broadcast on the $z$ dimension to get   $\hat{\boldsymbol{F}}_{xy}^{3D} \in \mathbb{R}^{X \times Y \times Z \times C}$. The broadcast features are denoted as $\hat{\boldsymbol{F}}^{TPV} = \{ \hat{\boldsymbol{F}}_{xy}^{3D}, \hat{\boldsymbol{F}}_{yz}^{3D}, \hat{\boldsymbol{F}}_{zx}^{3D}\}$. We fuse $\boldsymbol{F}^{3D}$ and $\hat{\boldsymbol{F}}^{TPV}$ by directly summing them and then apply a convolution layer followed by a Softmax function to generate the weight tensor $\boldsymbol{W}^{agg} \in \mathbb{R}^{X \times Y \times Z \times 4}$. In summary, the final voxel feature $\boldsymbol{F}^{final}$ is computed as:
\begin{align}
\label{fun:Wagg}
     \boldsymbol{W}^{Agg} &= \text{Softmax}(\text{Conv}(\boldsymbol{F}^{3D} + \sum{\hat{\boldsymbol{F}}^{TPV}})),\\
    \boldsymbol{F}^{final} &= \sum_{i}^{4}\boldsymbol{W}_i^{Agg}\odot \boldsymbol{F}^{3D}_i.
\end{align}

\PAR{SSC Head:} We use an SSC head consisting of linear projection layers and softmax to predict the final semantic occupancy $\boldsymbol{Y} \in \mathbb{R}^{X \times Y \times Z \times N_c}$, where $N_c$ denotes the number of target classes. Following~\cite{cao2022monoscene}, we train our network by minimizing the sum of three types of losses, including $\mathcal{L}_{ce}$, $\mathcal{L}_{scal}^{geo}$ and $\mathcal{L}_{scal}^{sem}$. The three losses represent cross-entropy loss weighted by class frequencies, and scene-class affinity loss for geometry and semantics respectively.

\subsection{Cross-Modal Knowledge Distillation} 
\label{KD}
We incorporate three types of distillation at different levels, including feature similarity distillation (FSD), TPV distillation (TPVD) and prediction alignment distillation (PAD), while FSD and TPVD focus on feature-level distillation and PAD focus on prediction-level distillation.

\PAR{Feature Similarity Distillation (FSD).} Utilizing the same occupancy prediction network for both the camera and LiDAR modalities enables feature similarity distillation. The primary objective is to develop a loss function that minimizes the discrepancies between feature maps at corresponding levels across different modalities. As illustrated in Figure \ref{fig:overrall}, we incorporate as many features as possible into the distillation process to achieve optimized outcomes. Considering the differences between the voxel feature extractors of the camera and LiDAR branches, we select cosine similarity loss as the loss function, as it exclusively focuses on the angular relationship between features while disregarding their magnitudes.

Let us denote the 2D and 3D features to distill by  $ \boldsymbol{F}^{2D}_{distill}=\{  \boldsymbol{f}_{i}^{2D} \in \mathbb{R}^{h\times w\times C} | i=0,\dots, N_{2d} \}$ and $ \boldsymbol{F}^{3D}_{distill}=\{  \boldsymbol{f}_{i}^{3D} \in \mathbb{R}^{X\times Y\times Z\times C} | i=0,\dots, N_{3d} \}$, respectively. $N_{2d}$ and $N_{3d}$ represent the number of 2D and 3D feature to be distilled.
To streamline the representation, we combine $ \boldsymbol{F}^{2D}_{distill}$ and $ \boldsymbol{F}^{3D}_{distill}$ into a single set denoted as $ \boldsymbol{F}_{distill}=\{  \boldsymbol{f}_{i} \in \mathbb{R}^{M\times C} | i=0,\dots, N=N_{2d}+N_{3D} \}$ where $M$ generaly refers to the number of elements in the 2D or 3D feature. Furthermore, assuming that $ \boldsymbol{f}_{i,T}(j)$ and $ \boldsymbol{f}_{i,S}(j)$ are the $j$-th element of the $i
$-th feature map from the teacher and student network, respectively. The feature similarity loss is formulated as follows:
\begin{align}
 \mathcal{L}_{fsd} = 1- \frac{1}{N}\sum^{N}_{i=1}\frac{1}{M}\sum^{M}_{j=1}\frac{ \boldsymbol{f}_{i,S}^\top(j) \boldsymbol{f}_{i,T}(j)}{|| \boldsymbol{f}_{i,S}(j)||_2|| \boldsymbol{f}_{i,T}(j)||_2}.
\end{align}
%

\begin{table*}
    \captionsetup{font={small}}
	\setlength{\tabcolsep}{0.0035\linewidth}
	\newcommand{\classfreq}[1]{{~\tiny(\semkitfreq{#1}\%)}}  %
	\centering
    \caption{Performance on SemanticKITTI\cite{behley2019semantickitti} (test set). The $\text{C},\text{L}$ denotes input modality of camera and LiDAR, respectively. We mark the best and the second best score in \textbf{bold} and \underline{underlined} and mark {\dag}  after the temporal stereo-based methods.
        }
   \resizebox{1\linewidth}{!}{
	\begin{tabular}{l |c |c c |c c c c c c c c c c c c c c c c c c c c}
 
		\toprule
		Method
		& \makecell[c]{Input}
		& \makecell[c]{IoU}
            & \makecell[c]{mIoU}
		& \rotatebox{90}{\textcolor{road}{$\blacksquare$} road}
		& \rotatebox{90}{\textcolor{sidewalk}{$\blacksquare$} sidewalk}
		& \rotatebox{90}{\textcolor{parking}{$\blacksquare$} parking} 
		& \rotatebox{90}{\textcolor{other-grnd.}{$\blacksquare$} other-grnd.} 
		& \rotatebox{90}{\textcolor{building}{$\blacksquare$} building} 
		& \rotatebox{90}{\textcolor{car}{$\blacksquare$} car} 
		& \rotatebox{90}{\textcolor{truck}{$\blacksquare$} truck} 
		& \rotatebox{90}{\textcolor{bicycle}{$\blacksquare$} bicycle} 
		& \rotatebox{90}{\textcolor{motorcycle}{$\blacksquare$} motorcycle} 
		& \rotatebox{90}{\textcolor{other-veh.}{$\blacksquare$} other-veh.} 
		& \rotatebox{90}{\textcolor{vegetation}{$\blacksquare$} vegetation} 
		& \rotatebox{90}{\textcolor{trunk}{$\blacksquare$} trunk} 
		& \rotatebox{90}{\textcolor{terrain}{$\blacksquare$} terrain} 
		& \rotatebox{90}{\textcolor{person}{$\blacksquare$} person} 
		& \rotatebox{90}{\textcolor{bicyclist}{$\blacksquare$} bicyclist} 
            & \rotatebox{90}{\textcolor{motorcyclist}{$\blacksquare$} motorcyclist}
            & \rotatebox{90}{\textcolor{fence}{$\blacksquare$} fence}
            & \rotatebox{90}{\textcolor{pole}{$\blacksquare$} pole}
		& \rotatebox{90}{\textcolor{traffic-sign}{$\blacksquare$} traffic-sign} \\

		\midrule
		  MonoScene~\cite{cao2022monoscene} & C & 34.16 & 11.08 & 54.70 & 27.10 & 24.80  & 5.70  &  14.40 &  18.80 & 3.30 & 0.50 &        0.70 & 4.40 & 14.90 & 2.40 & 19.50 & 1.00 & 1.40 & 0.40 & 11.10 & 3.30 & 2.10 \\
  
            TPVFormer~\cite{huang2023tri} & C & 34.25 & 11.26 & 55.10 & 27.20 & 27.40 & 6.50 & 14.80 & 19.20 & 3.70 & 1.00 & 0.50 & 2.30 & 13.90 & 2.60 & 20.40 & 1.10 & 2.40 & 0.30 & 11.00 & 2.90 & 1.50 \\
            
            OccFormer~\cite{zhang2023occformer} & C & 34.53 & 12.32 & 55.90 & 30.30 & 31.50 & 6.50 & 15.70 & 21.60 & 1.20 & 1.50 & 1.70 & 3.20 & 16.80 & 3.90 & 21.30 & 2.20 & 1.10 & 0.20 & 11.90 & 3.80 & 3.70 \\

            IAMSSC~\cite{10379527} & C & 43.74 & 12.37 & 54.00 & 25.50 & 24.70 & 6.90 & 19.20 & 21.30 & 3.80 & 1.10 & 0.60 & 3.90 & 22.70 & 5.80 & 19.40 & 1.50 & 2.90 & 0.50 & 11.90 & 5.30 & 4.10 \\
            
            DepthSSC~\cite{yao2023depthssc} & C & \underline{44.58} & 13.11 & 55.64 & 27.25 & 25.72 & 5.78 & 20.46 & 21.94 & 3.74 & 1.35 & 0.98 & 4.17 & 23.37 & 7.64 & 21.56 & 1.34 & 2.79 & 0.28 & 12.94 & 5.87 & 6.23 \\
            
            $\text{VoxFormer-T}^\dag$~\cite{li2023voxformer} & C & 43.21 & 13.41 & 54.10 & 26.90 & 25.10 & 7.30 & 23.50 & 21.70 & 3.60 & 1.90 & 1.60 & 4.10 & 24.40 & 8.10 & 24.20 & 1.60 & 1.10 & 0.00 & 13.10 & 6.60 & 5.70 \\

            $\text{HASSC-T}^\dag$~\cite{wang2024not} & C & 42.87 & 14.38 & 55.30 & 29.60 & 25.90 & 11.30 & 23.10 & 23.00 & 2.90 & 1.90 & 1.50 & 4.90 & 24.80 & 9.80 & 26.50 & 1.40 & 3.00 & 0.00 & 14.30 & 7.00 & 7.10 \\
            
            $\text{H2GFormer-T}^\dag$~\cite{wang2024h2gformer} & C & 43.52 & 14.60 & 57.90 & 30.40 & 30.00 & 6.90 & 24.00 & 23.70 & \underline{5.20} & 0.60 & 1.20 & 5.00 & \underline{25.20} & 10.70 & 25.80 & 1.10 & 0.10 & 0.00 & 14.60 & 7.50 & \underline{9.30} \\
            
            Symphonize~\cite{jiang2023symphonize} & C & 42.19 & 15.04 & 58.40 & 29.30 & 26.90 & 11.70 & 24.70 & 23.60 & 3.20 & 3.60 & 2.60 & 5.60 & 24.20 & 10.00 & 23.10 & \textbf{3.20} & 1.90 & \textbf{2.00} & 16.10 & 7.70 & 8.00 \\    
            
            StereoScene~\cite{li2023stereoscene} & C & 43.34 & 15.36 & 61.90 & 31.20 & 30.70 & 10.70 & 24.20 & 22.80 & 2.80 & 3.40 & 2.40 & 6.10 & 23.80 & 8.40 & 27.00 & \underline{2.90} & 2.20 & 0.50 & 16.50 & 7.00 & 7.20 \\

            MonoOcc-S~\cite{zheng2024monoocc} & C & - & 13.80 & 55.20 & 27.80 & 25.10 & 9.70 & 21.40 & 23.20 & \underline{5.20} & 2.20 & 1.50 & 5.40 & 24.00 & 8.70 & 23.00 & 1.70 & 2.00 & 0.20 & 13.40 & 5.80 & 6.40 \\
            MonoOcc-L~\cite{zheng2024monoocc} & C & - & 15.63 & 59.10 & 30.90 & 27.10 & 9.80 & 22.90 & 23.90 & \textbf{7.20} & \underline{4.50} & 2.40 & \textbf{7.70} & 25.00 & 9.80 & 26.10 & 2.80 & \textbf{4.70} & \underline{0.60} & 16.90 & 7.30 & 8.40 \\

            CGFormer~\cite{yu2024context} & C & 44.41 & 16.63 & 64.30 & 34.20 & \underline{34.10} &12.10 & \textbf{25.80} & 26.10 & 4.30 & 3.70 & 1.30 & 2.70 & 24.50 & 11.20 & 29.30 & 1.70 & \underline{3.60} & 0.40 & 18.70 & 8.70 & \underline{9.30} \\

            \midrule
            
            {{\methodname}-C} (Ours) & C & 44.31 & \underline{17.03} & \underline{66.00} & \underline{35.00} & 33.10 & \underline{13.50} & \underline{25.10} & \underline{27.20} & 3.00 & 3.50 & \textbf{3.60} & 4.30 & \underline{25.20} & \underline{11.50} & \underline{30.10} & 1.50 & 2.40 & 0.20 & \underline{20.50} & \underline{9.10} & 8.90   \\
            {{\methodname}-D} (Ours) & C & \textbf{45.37} & \textbf{18.18} & \textbf{68.20} & \textbf{36.90} & \textbf{34.60} & \textbf{16.20} & \textbf{25.80} & \textbf{28.30} & 4.50 & \textbf{4.90} & \underline{3.30} & \underline{7.20} & \textbf{26.20} & \textbf{11.90} & \textbf{32.00} & 2.10 & 2.40 & 0.30 & \textbf{21.60} & \textbf{9.60} & \textbf{9.50}   \\
            
            \midrule
            {{\methodname}-L} (Ours) & L & 60.32 & 23.37 & 68.50 & 40.60 & 33.20 & 6.10 & 41.50 & 36.80 & 5.40 & 8.70 & 4.10 & 9.00 & 42.60 & 28.70 & 36.90 & 1.40 & 2.90 & 1.00 & 27.70 & 27.00 & 21.90   \\
        \bottomrule
	\end{tabular}} \\
	\label{table:semkitti_main}
\end{table*}
\begin{table*}
    \captionsetup{font={small}}
	\setlength{\tabcolsep}{0.0035\linewidth}
	\newcommand{\classfreq}[1]{{~\tiny(\semkitfreq{#1}\%)}}  %
	\centering
    \caption{Performance on SSCBench-KITTI360\cite{li2023sscbench} (test set). We also mark the best and the second best score in \textbf{bold} and \underline{underlined}.
    }
   \resizebox{1\linewidth}{!}{
	\begin{tabular}{l |c |c c |c c c c c c c c c c c c c c c c c c c c}
 
		\toprule
		Method
		& \makecell[c]{Input}
		& \makecell[c]{IoU}
            & \makecell[c]{mIoU}
		& \rotatebox{90}{\textcolor{car}{$\blacksquare$} car}
		& \rotatebox{90}{\textcolor{bicycle}{$\blacksquare$} bicycle}
		& \rotatebox{90}{\textcolor{motorcycle}{$\blacksquare$} motorcycle} 
		& \rotatebox{90}{\textcolor{truck}{$\blacksquare$} truck} 
		& \rotatebox{90}{\textcolor{other-veh.}{$\blacksquare$} other-veh.} 
		& \rotatebox{90}{\textcolor{person}{$\blacksquare$} person} 
		& \rotatebox{90}{\textcolor{road}{$\blacksquare$} road} 
		& \rotatebox{90}{\textcolor{parking}{$\blacksquare$} parking} 
		& \rotatebox{90}{\textcolor{sidewalk}{$\blacksquare$} sidewalk} 
		& \rotatebox{90}{\textcolor{other-grnd.}{$\blacksquare$} other-grnd.} 
		& \rotatebox{90}{\textcolor{building}{$\blacksquare$} building} 
		& \rotatebox{90}{\textcolor{fence}{$\blacksquare$} fence} 
		& \rotatebox{90}{\textcolor{vegetation}{$\blacksquare$} vegetation} 
		& \rotatebox{90}{\textcolor{terrain}{$\blacksquare$} terrain} 
		& \rotatebox{90}{\textcolor{pole}{$\blacksquare$} pole} 
        & \rotatebox{90}{\textcolor{traffic-sign}{$\blacksquare$} traffic-sign}
        & \rotatebox{90}{\textcolor{other-struct.}{$\blacksquare$} other-struct.}
        & \rotatebox{90}{\textcolor{other-obj.}{$\blacksquare$} other-obj.} \\

		\midrule
		  MonoScene~\cite{cao2022monoscene} & C & 37.87 & 12.31 & 19.34 & 0.43 & 0.58 & 8.02 & 2.03 & 0.86 & 48.35 & 11.38 & 28.13 & 3.32 & 32.89 & 3.53 & 26.15 & 16.75 & 6.92 & 5.67 & 4.20 & 3.09 \\
    
            TPVFormer~\cite{huang2023tri} & C & 40.22 & 13.64 & 21.56 & 1.09 & 1.37 & 8.06 & 2.57 & 2.38 & 52.99 & 11.99 & 31.07 & 3.78 & 34.83 & 4.80 & 30.08 & 17.52 & 7.46 & 5.86 & 5.48 & 2.70 \\

            OccFormer~\cite{zhang2023occformer} & C & 40.27 & 13.81 & 22.58 & 0.66 & 0.29 & 9.89 & 3.82 & 2.77 & 54.30 & 13.44 & 31.53 & 3.55 & 36.42 & 4.80 & 31.00 & 19.51 & 7.77 & 8.51 & 6.95 & 4.60\\

            VoxFormer~\cite{li2023voxformer} & C & 38.76 & 11.91 & 17.84 & 1.16 & 0.89 & 4.56 & 2.06 & 1.63 & 47.01 & 9.67 & 27.21 & 2.89 & 31.18 & 4.97 & 28.99 & 14.69 & 6.51 & 6.92 & 3.79 & 2.43\\

            IAMSSC~\cite{10379527} & C & 41.80 & 12.97 & 18.53 & 2.45 & 1.76 & 5.12 & 3.92 & 3.09 & 47.55 & 10.56 & 28.35 & 4.12 & 31.53 & 6.28 & 29.17 & 15.24 & 8.29 & 7.01 & 6.35 & 4.19 \\

            DepthSSC~\cite{yao2023depthssc} & C & 40.85 & 14.28 & 21.90 & 2.36 & 4.30 & 11.51 & 4.56 & 2.92 & 50.88 & 12.89 & 30.27 & 2.49 & 37.33 & 5.22 & 29.61 & 21.59 & 5.97 & 7.71 & 5.24 & 3.51 \\

            Symphonies~\cite{jiang2023symphonize} & C & 44.12 & 18.58 & \underline{30.02} & 1.85 & \textbf{5.90} & \textbf{25.07} & \textbf{12.06} & \textbf{8.20} & 54.94 & 13.83 & 32.76 & \textbf{6.93} & 35.11 & \underline{8.58} & 38.33 & 11.52 & 14.01 & 9.57 & \textbf{14.44} & \textbf{11.28} \\

            CGFormer~\cite{yu2024context} & C & \underline{48.07} & 20.05 & 29.85 & 3.42 & 3.96 & \underline{17.59} & 6.79 & 6.63 & \underline{63.85} & 17.15 & \underline{40.72} & 5.53 & 42.73 & 8.22 & 38.80 & \underline{24.94} & \underline{16.24} & 17.45 & 10.18 & 6.77 \\

            \midrule
            
            {{\methodname}-C} (Ours) & C & \underline{48.07} & \underline{20.11} & 29.62 & \textbf{3.71} & 4.35 & 14.93 & 8.35 & \underline{7.23} & 63.31 & \underline{17.91} & 40.51 & 5.23 & \underline{42.78} & 8.52 & \underline{39.35} & 24.52 & 16.22 & \underline{18.44} & 10.24 & 6.79   \\
            
            {{\methodname}-D} (Ours) & C & \textbf{48.83} & \textbf{20.99} & \textbf{30.57} & \underline{3.70} & \underline{4.63} & 17.34 & \underline{9.73} & 7.00 & \textbf{64.19} & \textbf{18.35} & \textbf{41.24} & \underline{6.42} & \textbf{43.91} & \textbf{9.72} & \textbf{39.88} & \textbf{26.02} & \textbf{16.26} & \textbf{19.40} & \underline{10.65} & \underline{8.87}   \\
            
            \midrule
            {{\methodname}-L} (Ours) & L & 57.60 & 25.22 & 40.41 & 4.08 & 4.23 & 26.15 & 10.30 & 9.94 & 70.01 & 22.62 & 46.16 & 8.78 & 51.04 & 14.34 & 47.60 & 31.33 & 24.89 & 21.64 & 13.96 & 6.41  \\
        \bottomrule
	\end{tabular}} \\
	\label{table:kitti360_main}
\end{table*}

\PAR{TPV Distillation (TPVD):}
We have both the feature relation and aggregator distillation at the TPV feature level.
\subsubsection{TPV Relation Distillation (TRD)}
 Inspired by~\cite{ma2024licroccteachradaraccurate}, we introduce TPV relation distillation, emphasizing the consistency of the relations among TPV features from both the teacher and student models. For instance, considering the XY plane, teacher feature $ \boldsymbol{\hat{F}}_{xy,t}$ and student feature $ \boldsymbol{\hat{F}}_{xy, s} \in \mathbb{R}^{X \times Y \times C}$ are initially generated from the TPV encoder, we subsequently adjust the dimension of these tensors to fit within the space of $ \boldsymbol{\mathbb{R}}^{C \times (X \times Y)}$ and compute affinity matrix as follows:
\begin{align}
    {C}(u,v) = \frac{ \boldsymbol{f}_{u}^\top  \boldsymbol{f}_{v}}{|| \boldsymbol{f}_{u}||_2|| \boldsymbol{f}_{v}||_2},(u,v \in \{ 1,2,\dots ,K=X \times Y\}),
\end{align}
where ${C}(u,v)$ denotes the cosine similarity at each element $(u, v)$ of the affinity matrix, $ \boldsymbol{f}_{u}$ denotes the $u$-th feature in the feature map $ \boldsymbol{F}_{xy}$. The TRD loss on the XY plane is then defined as follows:
\begin{align}
  \mathcal{L}_{trd}^{xy} = \frac{1}{K\times K}\sum^{K}_{u=1}\sum^{K}_{v=1}||{C}^{xy}_{S}(u,v)-{C}^{xy}_{T}(u,v)||_1,
\end{align}
where $\boldsymbol{C}^{xy}_{s}$ and $\boldsymbol{C}^{xy}_{t}$ denote the affinity matrix of the student and teacher network's XY plane TPV feature maps, respectively, while $\mathcal{L}_{trd}^{xy}$ quantifies the discrepancy between these matrices. The final TRD loss is computed as the sum of the contributions from three planes:
\begin{align}
  \mathcal{L}_{trd} = \mathcal{L}_{trd}^{xy} + \mathcal{L}_{trd}^{yz} + \mathcal{L}_{trd}^{zx}
\end{align}
\subsubsection{TPV Aggregator Distillation (TAD)}
TPV aggregator distillation directly aligns the student's weight distribution with the teacher's, thereby facilitating the student's convergence towards the teacher's parameters. Inspired by Monoscene~\cite{cao2022monoscene}, we employ KL Divergence~\cite{kullback1951information} loss to measure the difference in aggregating weight distribution, assuming that $\boldsymbol{W}^{Agg}_S$ and $\boldsymbol{W}^{Agg}_T$, calculated by Equation~\ref{fun:Wagg}, represent the weight tensors from the student and teacher models, respectively, TPV aggregator distillation loss, denoted as $\mathcal{L}_{tad}$ is computed as follows:
\begin{align}
    \mathcal{L}_{tad} = \text{KL}(\boldsymbol{W}^{Agg}_S||\boldsymbol{W}^{Agg}_T)
\end{align}

\PAR{Prediction Alignment Distillation (PAD):}
Prediction alignment distillation seeks to align the student and teacher models by minimizing the distributional discrepancies. Like TPV aggregator distillation, we also use KL Divergence loss to measure the difference in distribution between predicted targets. The PAD loss is computed as follows:
\begin{align}
 \mathcal{L}_{pad} = \text{KL}(\boldsymbol{\widetilde{Y}}_S||\boldsymbol{\widetilde{Y}}_T)
\end{align}

\subsection{Training Loss}
\label{SML}
In this section, we will review all the loss functions previously mentioned and summarize them into an overall loss function. To optimize the Depth Net in the view transformer, we incorporate $\mathcal{L}_{d}$ (Section~\ref{voxelfeature}). For occupancy prediction network, we employ $\mathcal{L}_{ce}$, $\mathcal{L}_{scal}^{geo}$, $\mathcal{L}_{scal}^{sem}$ (Section~\ref{occnet}). To enable the knowledge distillation, we take $\mathcal{L}_{fsd}$, $\mathcal{L}_{trd}$, $\mathcal{L}_{tad}$ and $\mathcal{L}_{pad}$ into use (Section~\ref{KD}). The overall loss function is formulated as follows:
\begin{multline}
    \mathcal{L} = \lambda_1\mathcal{L}_{ce} + \lambda_2\mathcal{L}_{scal}^{geo} + \lambda_3\mathcal{L}_{scal}^{sem} + \lambda_4\mathcal{L}_{d} \\
    +\lambda_5\mathcal{L}_{fsd} + \lambda_6\mathcal{L}_{trd} + \lambda_7\mathcal{L}_{tad} + \lambda_8\mathcal{L}_{pad}
\end{multline}
where $\lambda_1$ to $\lambda_8$ are hyper-parameters. We set the values of these hyper paramters to be 3, 1.5, 0.5, 0.001, 4, 5, 10, 70, respectively, in our experiments.

\section{EXPERIMENTS}
\label{sec:experiement}

\subsection{Datasets And Metrics}
We evaluate our algorithm on the SemanticKITTI~\cite{behley2019semantickitti} and SSCBench-KITTI360 datasets~\cite{li2023sscbench}. Both datasets provide heterogeneous sensor measurements, including stereo RGB images and LiDAR scans. Dense semantic occupancy annotations are available for 22 and 9 urban driving sequences, respectively, covering the range of $51.2m \times 51.2m \times 6.4m$, represented as voxel grids of $256 \times 256 \times 32$, with a voxel size of $0.2 \, \text{m}$. 
We report the commonly-used metrics for SSC tasks, including mIoU and IoU regarding the accuracy, along with memory usage and inference time for efficiency.

\begin{table}
    \captionsetup{font={small}}
        \caption{Memory usage and inference time tested on the SemanticKITTI dataset. In alignment with \cite{yu2024context} , these metrics are measured on the NVIDIA RTX 4090 GPU.
        }
	\centering\resizebox{1\linewidth}{!}
        {
	\begin{tabular}{l |c |c c c }
 
		\toprule
		Method
		& \makecell[c]{mIoU}
		& \makecell[c]{Train. Mem.}
            & \makecell[c]{Infer. Mem.}
            & \makecell[c]{Infer. Time}
		  \\

		\midrule
            Symphonize~\cite{jiang2023symphonize} & 15.04 & 17757M & 6360M & 216ms  \\
            
            StereoScene~\cite{li2023stereoscene} & 15.36 & 19000M & 7329M & 258ms  \\

            MonoOcc-S~\cite{zheng2024monoocc} & 13.80 & 18432M & 8192M &  295ms \\

            CGFormer~\cite{yu2024context} & 16.63 & 19330M & 6550M & 205ms   \\

            \midrule
            
            {{\methodname}-C} (Ours) & \textbf{17.03} & \textbf{13068M} & \textbf{5012M} & \textbf{151ms}    \\
            {{\methodname}-D} (Ours) & \textbf{18.18} & \textbf{14528M} & \textbf{5012M} & \textbf{151ms}   \\

        \bottomrule
	\end{tabular}} \\

	\label{table:semkitti_efficiency}
\end{table}
\begin{figure*}[t]
      \centering
      \includegraphics[width=\linewidth]{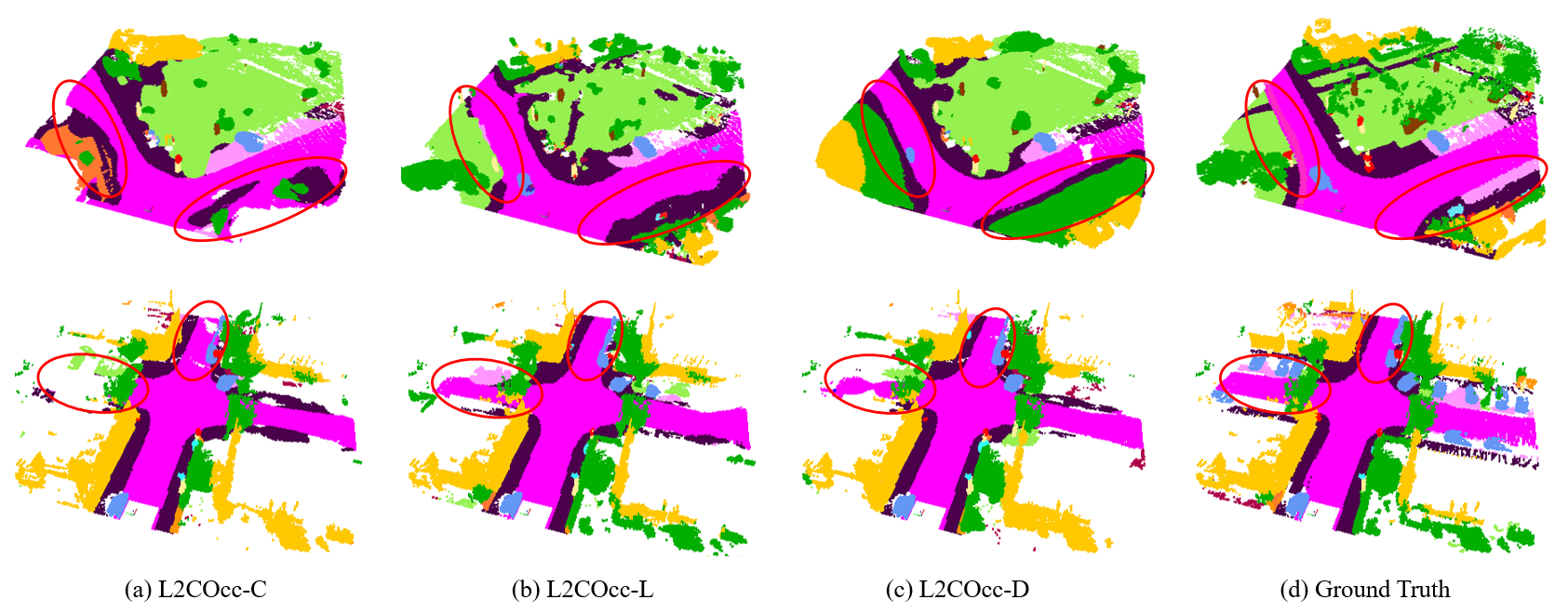}
      \vspace{-10pt}
      \captionsetup{font={small}}
      \caption{Qualitative visualizations on SemanticKITTI and SSCBench-KITTI360. The introduction of cross-modal knowledge distillation enables L2COcc-D to achieve superior performance compared to L2COcc-C, particularly in regions farther from the camera and at the boundaries of the camera's field of view.
      }
      \label{fig:ssc_vis}
      \vspace{-5pt} 
\end{figure*}
\subsection{Implementation Details}
We train our {\methodname} for 25 epochs on NVIDIA RTX 4090 GPUs, using batch sizes of 4 and 8 for SemanticKITTI~\cite{behley2019semantickitti} and SSCBench-KITTI360~\cite{li2023sscbench} benchmarks, respectively. We adopt AdamW optimizer, initialized with the maximum learning rate of 2e-4 and weight decay of 1e-2. The learning rate is adjusted according to a cosine annealing strategy.

\subsection{Quantitative Results}
\label{evaluation on semantickitti}
We conduct comparative experiments on both SemanticKITTI and SSCBench-KITTI360 datasets. 
We concisely denote our camera-based model and LiDAR-based model as {\methodname-C} and {\methodname-L} respectively, and refer to the camera-based model employing cross-modal knowledge distillation as {\methodname-D}.
Table~\ref{table:semkitti_main} and~\ref{table:semkitti_efficiency} present a comparison of our {\methodname} and the  state-of-the-art methods (SOTA) on the SemanticKITTI dataset. 
We only compare our {\methodname} with camera-based methods, as {\methodname-C} and {\methodname-D} require camera input only during inference.
As demonstrated in Table~\ref{table:semkitti_main}, our visual-base methods {\methodname-D} and {\methodname-C} achieve mIoU of 18.18 and 17.03, ranking first and second in semantic prediction,
respectively.  Cross-modal knowledge distillation significantly enhances accuracy across nearly all categories. Table~\ref{table:semkitti_efficiency} indicates that {\methodname} completes scene inference in 151 milliseconds and utilizes memory less than 5GB GPU, leading to over a 23\% savings in both memory usage and inference time compared to the second-best method. 
Simultaneously, the reduction in GPU memory usage during training enables L2COcc to be trained on a more cost-effective computing platform. 
Moreover, we achieve SOTA performance on the SSC-KITTI360 dataset (Table \ref{table:kitti360_main}), further demonstrating the excellence of our {\methodname}.

\subsection{Ablation Studies}
To validate the efficacy of our proposed module, we conduct a series of ablation experiments on the validation set of SemanticKITTI.

\PAR{Ablation on Efficient Voxel Transformer (EVT).}
This section aims to elucidate the rationale and effectiveness of the enhancements made to the voxel feature extraction process (see Sec.~\ref{voxelfeature}).
In contrast to CGFormer~\cite{yu2024context} that integrates the LSS volume with learnable voxel embeddings as voxel queries, subsequently generating voxel features through 3D-DCA and 3D-DSA modules, we directly use the LSS volume as voxel queries and replace 3D-DSA module with a lightweight 3D encoder. The findings in Table \ref{table:ablation_1} indicate that the 3D-DSA module consumes up to 6G of GPU memory, yet it leads to no significant performance enhancements. Additionally, the voxel embedding introduces 34 million extra parameters but maintains almost the same mIoU as the version without voxel embedding. The 4-th row of Table \ref{table:ablation_1} demonstrates that our lightweight 3D encoder is effective in enhancing the model's performance.

\begin{table}[H]
    \captionsetup{font={small}}
	\caption{Ablation on efficient voxel transformer.
        }
	\centering\resizebox{1\linewidth}{!}
        {
	\begin{tabular}{l |c c c |c c c }
 
		\toprule
		row
		& \makecell[c]{3D-DSA}
            & \makecell[c]{\makecell{Voxel \\ Embeding}}
            & \makecell[c]{\makecell{3D \\ Encoder}}
            & \makecell[c]{mIoU}
            & \makecell[c]{Parameters}
            & \makecell[c]{\makecell{Train \\ Mem.}}
		  \\

		\midrule
            1 & \cmark & \cmark & \xmark & 15.86 & 148M & 17448M \\
            
            2 & \xmark & \cmark & \xmark & 16.34 & 148M & 11326M \\

            3 & \xmark & \xmark & \xmark & 16.40 & 114M & 10849M \\

            4 & \xmark & \xmark & \cmark & 16.72 & 119M & 13068M \\

        \bottomrule
	\end{tabular}} \\
	\label{table:ablation_1}
\end{table}
\begin{figure}[t]
      \centering
      \includegraphics[width=\linewidth]{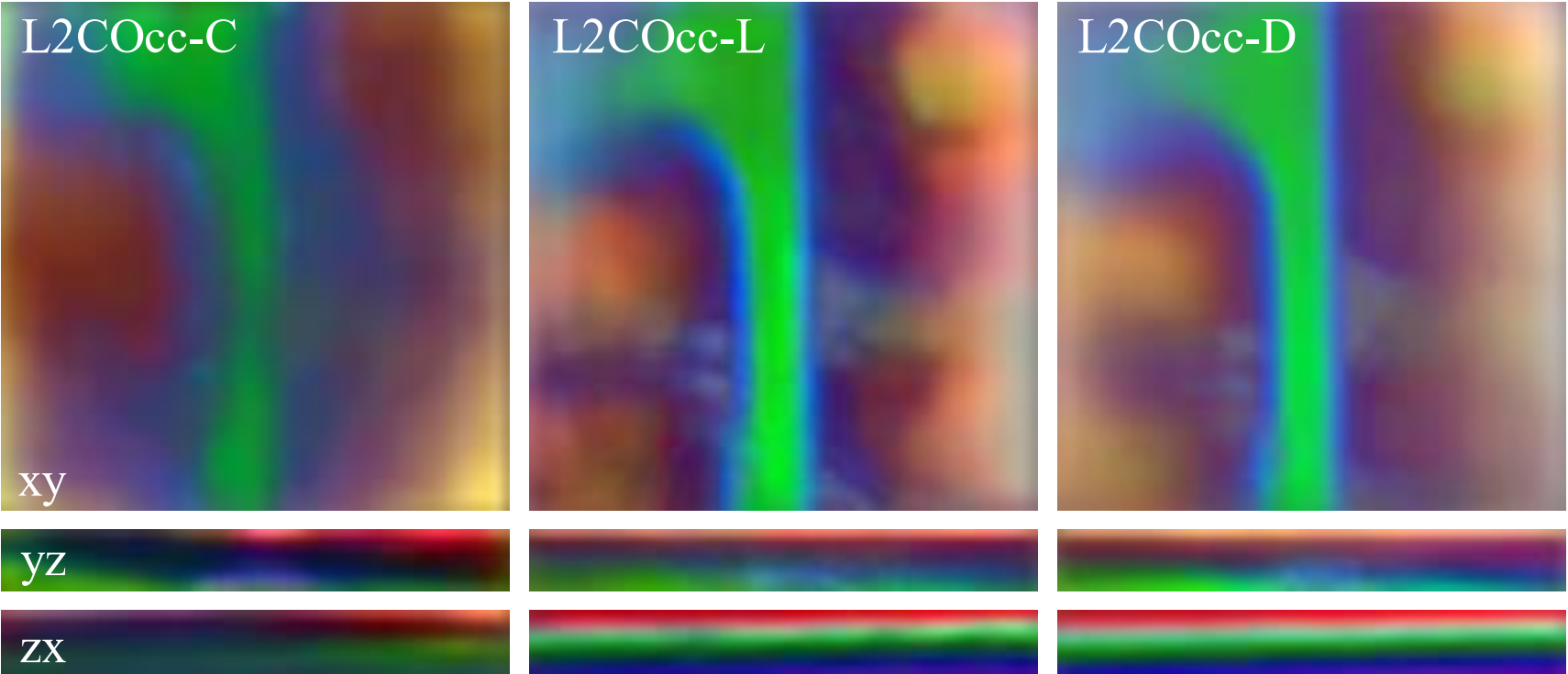}
      \captionsetup{font=small}
      \caption{PCA visualization results of TPV features $\boldsymbol{F^{Enc}}$. 
      $\boldsymbol{F^{Enc}}$ acquired from LiDAR exhibit a distinctive and well-defined scene geometry, whereas $\boldsymbol{F^{Enc}}$ acquired from the camera display a blurred and radial appearance. The introduction of knowledge distillation effectively addresses this challenge. More qualitative results are provided at our project page: \url{https://studyingfufu.github.io/L2COcc/}.}
      \label{fig:tpv}
      \vspace{-10pt} 
\end{figure}

\PAR{Ablation on TPV Processing Modules.} 
The outcomes presented in Table~\ref{table:ablation_2} underscore the effectiveness of our proposed TPV pooler and TPV aggregator (see Sec.~\ref{occnet}) by comparing them with the TPV processing modules utilized in CGFormer. 
GMP denotes the Grouped Max Pooling and WAP denotes the Weighted Average Pooling we proposed. $\text{Agg}_{\text{CGFormer}}$ and $\text{Agg}_{\text{\methodname}}$ denotes the operation aggregating TPV feature proposed by CGFormer and our \methodname.
It is shown that substituting GMP with WAP results in a 0.34 mIoU improvement. Furthermore, refinements to the TPV aggregator contributed to an additional 0.54 mIoU enhancement.

\begin{table}[H]
    \captionsetup{font={small}}
	\caption{Ablation on TPV processing modules. }
	\centering\resizebox{0.75\linewidth}{!}
        {
	\begin{tabular}{l |c c |c }
 
		\toprule
		row
            & \makecell[c]{\makecell{TPV Pooler}}
            & \makecell[c]{\makecell{TPV Aggregator}}
            & \makecell[c]{mIoU}
		  \\

		\midrule
            
            2 & GMP & $\text{Agg}_{\text{CGFormer}}$ & 15.84 \\

            3 & WAP & $\text{Agg}_{\text{CGFormer}}$ &  16.18   \\

            4 & WAP & $\text{Agg}_{\text{\methodname}}$ & 16.72 \\

        \bottomrule
	\end{tabular}} \\
	\label{table:ablation_2}
\end{table}

\begin{figure}[t]
      \centering
      \includegraphics[width=\linewidth]{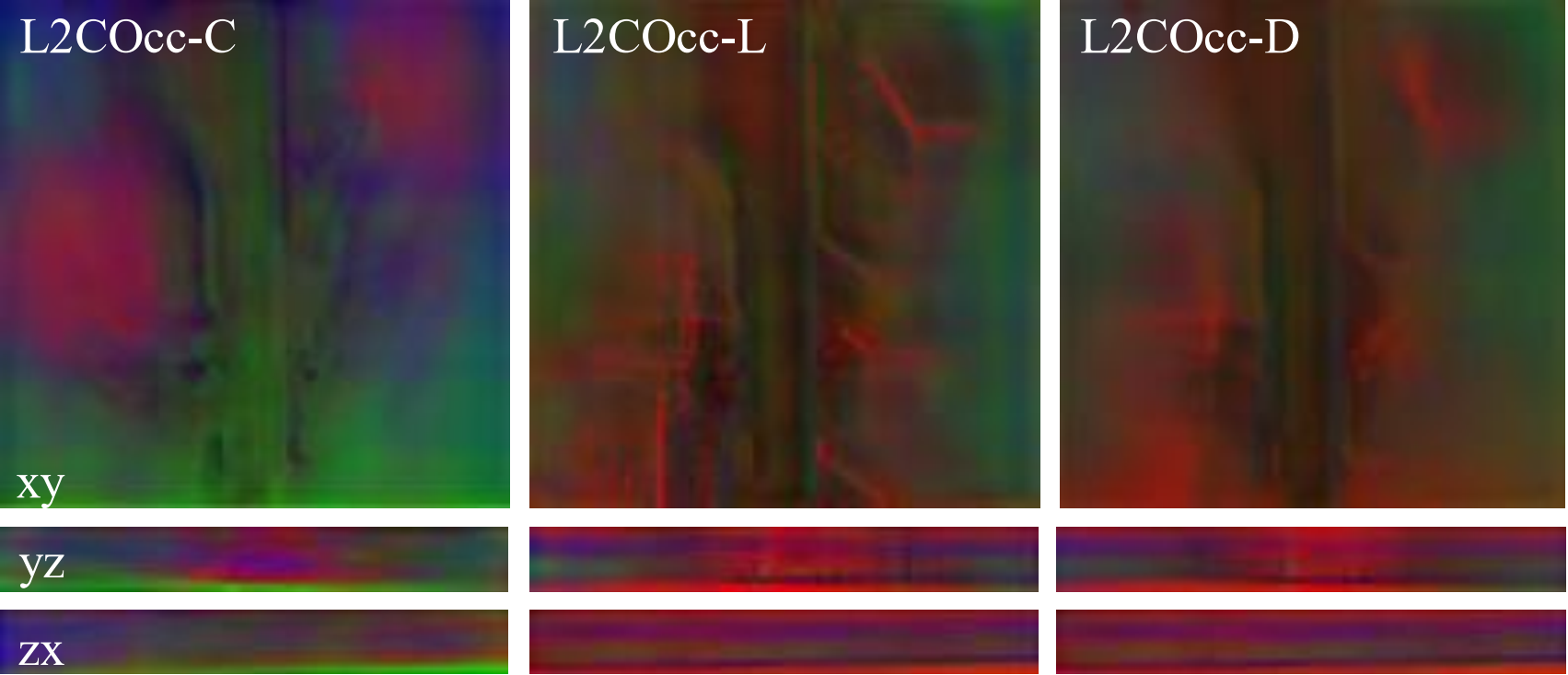}
      \captionsetup{font=small}
      \caption{Visualization results of TPV aggregating weight $\boldsymbol{W^{Agg}}$. We compute the average of the 3D weights across each dimension to generate three corresponding 2D weight maps. Subsequently, the weights associated with the TPV features' XY, YZ, and ZX planes are mapped to the R, G, and B channels, respectively. It's shown that the camera-based model emphasizes the YZ plane more, whereas the LiDAR-based model assigns more weight to the XY plane. Incorporating TPV aggregator distillation enables the camera model to adopt the weight allocation pattern learned by the LiDAR model.}
      \label{fig:agg}
      \vspace{-10pt} 
\end{figure}

\PAR{Ablation on Feature-Level Distillation (FSD,TPVD).}
FSD and TPVD aim to minimize feature discrepancies between the teacher and student models, thus referred to as feature-level distillation (see Sec.~\ref{KD}). TRD and PAD exhibit a more pronounced effect only when applied in tandem, and thus, we report only the combined results. The first three rows of Table \ref{table:ablation_3} indicate a progressive increase in mIoU from 16.72 to 17.25 with the integration of the FSD and TPVD modules. The qualitative findings presented in Figures \ref{fig:tpv} and \ref{fig:agg} demonstrate that the FSD and TPVD modules promote the convergence of the student model's TPV features and TPV aggregation weights to align with those of the teacher model.

\begin{table}[H]
    \captionsetup{font={small}}
	\caption{Ablation on Cross-Modal Knowledge Distillation.}
        \centering
	\resizebox{0.55\linewidth}{!}
        {
	\begin{tabular}{l |c c c |c }
 
		\toprule
		row
		& \makecell[c]{FSD}
            & \makecell[c]{TPVD}
            & \makecell[c]{PAD}
            & \makecell[c]{mIoU}

		  \\

		\midrule
            1 & \xmark & \xmark & \xmark & 16.72 \\
            
            2 & \cmark & \xmark & \xmark & 16.99 \\
            
            3 & \cmark & \cmark & \xmark & 17.25 \\

            4 & \xmark & \xmark & \cmark & 17.88 \\

            5 & \cmark & \cmark & \cmark & 18.22 \\

        \bottomrule
	\end{tabular}} \\
	\label{table:ablation_3}
\end{table}
\PAR{Ablation on Prediction Alignment Distillation (PAD).}
The PAD module directly affects the network's output to ensure alignment between the student network's prediction distribution and the teacher model's (see Sec.~\ref{KD}). 
The penultimate and final rows of Table \ref{table:ablation_3} illustrate the impact of employing the PAD module independently and in tandem with FSD and TPVD. 
The findings reveal that the standalone use of PAD can improve the mIoU by 1.16, with enhanced performance observed when integrated with FSD and TPVD. The combined application of both techniques results in a mIoU boost of up to 1.5. The visualization results in \ref{fig:ssc_vis} further demonstrate the effectiveness of cross-modal knowledge distillation.
\section{Conclusion}
\label{sec:conclusion}
This paper introduces {\methodname}, a lightweight camera-centric SSC network that also supports LiDAR input. We propose an efficient voxel transformer, and develop a TPV-based occupancy prediction pipeline for accurate and efficient SSC. Promising performance is attained by applying knowledge distillation from the LiDAR-based model to the camera-based model. Experiments show our {\methodname} achieves state-of-the-art results on SemanticKITTI and SSCBench-KITTI360 while consuming significantly less memory and computation time. 
For future work, it is worth investigating attentive distillation from the LiDAR model to the camera model, to accommodate the performance variations of models in different scenarios.






{
\AtNextBibliography{\small}
\printbibliography

}

\end{document}